\documentclass[letterpaper, 10 pt, conference]{ieeeconf}  % Comment this line out if you need a4paper

\usepackage[colorinlistoftodos]{todonotes}
\usepackage{cite}
\usepackage{booktabs}

\IEEEoverridecommandlockouts                              % This command is only needed if 
\usepackage{amsmath} % assumes amsmath package installed
\usepackage{amssymb}  % assumes amsmath package installed
\usepackage{comment}
\usepackage{multirow}
\usepackage{makecell}

\title{\LARGE \bf
EgoReveal: Learning Active Vision and Object Manipulation under Partial Occlusions from Egocentric Demonstrations
}

\author{Hiromu Kurihara$^{1}$, Kazutoshi Tanaka$^{2}$, Masashi Hatano$^{1}$, Tadashi Kozuno$^{2}$, Hideo Saito$^{1}$, Mariko Isogawa$^{1}$% <-this % stops a space
\thanks{*This work was supported by JSPS KAKENHI Grant Number JP25H01159.}% <-this % stops a space
\thanks{$^{1}$Keio University, Yokohama, Kanagawa 223-8522, Japan
        {\tt\small \{chestnuts36015, mariko.isogawa\}@keio.jp}}%
\thanks{$^{2}$OMRON SINIC X Corporation, Bunkyo-ku, Tokyo 113-0033, Japan
        {\tt\small kazutoshi.tanaka@sinicx.com}}%
}

\begin{document}

\maketitle
\thispagestyle{empty}
\pagestyle{empty}

%%%%%%%%%%%%%%%%%%%%%%%%%%%%%%%%%%%%%%%%%%%%%%%%%%%%%%%%%%%%%%%%%%%%%%%%%%%%%%%%

\begin{abstract}
Learning from Demonstrations (LfD) enables robots to learn object manipulation directly from human demonstrations.
In contrast to resource-intensive and labor-intensive teleoperation, egocentric demonstrations are easy to collect, enabling scalable data collection.
Manipulated objects are often partially occluded in everyday environments, yet prior egocentric demonstration-based methods that require no robots or specialized teaching devices do not explicitly address occlusions.
To fill this gap, we propose EgoReveal, a framework that learns viewpoint-repositioning together with manipulation actions from egocentric human demonstrations to reposition the viewpoint and obtain target visibility under partial occlusions.
EgoReveal instantiates this as two imitation policies: one predicts the future camera-arm pose sequence from the current view and camera-arm pose, and the other predicts the future control commands from object keypoints and a history of control commands.
EgoReveal switches between the two based on target visibility.
In experiments where the manipulated object is partially occluded initially, EgoReveal repositions the camera to obtain visibility and completes the task more reliably than fixed-camera baselines.
% TODO: add results
\end{abstract}

% TODO: non-learning active vision 
\section{INTRODUCTION}
Learning from Demonstrations (LfD), such as imitation learning (IL), enables robots to learn object manipulation directly from human demonstrations~\cite{zhangDeepImitationLearning2018}.
In LfD via teleoperation, a human operator remotely controls the robot while the system records the robot's observations and actions, yielding directly usable action labels without requiring any mapping from the human body to the robot~\cite{zhaoLearningFineGrainedBimanual2023a,chiDiffusionPolicyVisuomotor2025}.
Despite this advantage, teleoperation is resource-intensive and labor-intensive, since it requires specialized hardware and skilled operators, thereby limiting scalability to diverse, large-scale datasets.
Egocentric demonstrations are captured as first-person videos using an egocentric camera (e.g., smart glasses).
In contrast to teleoperation, egocentric demonstrations are easy to collect, enabling scalable data collection~\cite{liuEgoZeroRobotLearning2025a, guzeyDexteritySmartLenses2025a}.

Manipulated objects are often partially occluded in everyday environments by the hands, surrounding objects, or clutter, obscuring task-relevant visual cues.
In this study, we focus on the common case where the target remains in view but is partially occluded at the start of the episode.
Egocentric demonstrations capture naturally changing viewpoints, with humans adjusting their viewpoint when targets become hard to see, yet many egocentric demonstration-based methods assume a fixed camera viewpoint at inference time~\cite{liuEgoZeroRobotLearning2025a, lepertPhantomTrainingRobots2025a}.
This fixed-view assumption can cause the robot to fail the task by preventing access to task-relevant visual cues under occlusions.

\begin{figure}[t]
    \centering
    \includegraphics[width=1.0\linewidth]{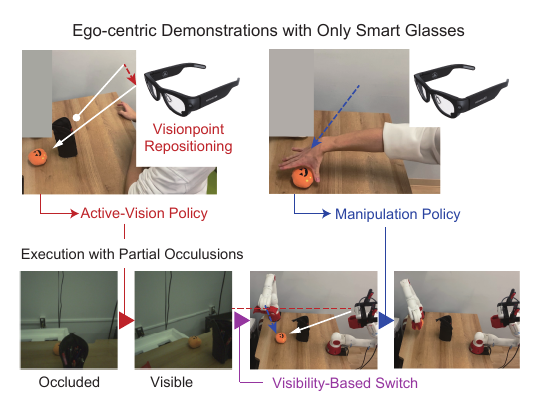}
    \caption{
    Conceptual illustration of EgoReveal.
    EgoReveal is a framework that learns viewpoint-repositioning to obtain target visibility under partial occlusions together with object manipulation from egocentric human demonstrations.
    }
    \label{fig:teaser}
\end{figure}

To ensure access to task-relevant visual cues under occlusions, recent studies use Head-Mounted Displays (HMDs) and enable active vision by learning to move a robot-mounted camera from an operator's viewpoint changes~\cite{chuangActiveVisionMight2025b, xiongVisionActionLearning2025a}.
These studies also rely on teleoperation or specialized teaching devices, increasing hardware costs and operational burden.

To fill this gap, we propose EgoReveal, a framework that learns viewpoint-repositioning together with manipulation actions from egocentric human demonstrations to reposition the viewpoint and obtain target visibility under partial occlusions.
EgoReveal instantiates this as two imitation policies: one predicts camera-arm pose trajectories from the current view and camera-arm state, and the other predicts end-effector actions from keypoints tracked in the actively moved camera view. 
EgoReveal switches between the two based on target visibility.
In experiments where the manipulated object is partially occluded initially, EgoReveal repositions the camera to obtain visibility and completes the task more reliably than fixed-camera baselines.
L% The primary contribution of this paper is a learning framework that uses natural egocentric human demonstrations, collected without teleoperation or specialized teaching devices, to learn viewpoint-repositioning alongside manipulation actions to handle partial occlusions.

The contributions of this paper are twofold:
\begin{itemize}
    \item Formulation: A problem setting for learning manipulation from egocentric human demonstrations without robotic teleoperation or handheld teaching devices, under partial occlusions.
    \item EgoReveal: A framework for learning manipulation from egocentric human demonstrations that couples the Active Vision policy and the Manipulation policy via visibility-based switching to handle partial occlusions.
\end{itemize}

\section{RELATED WORK}
This section summarizes related work on LfD from egocentric demonstrations and learning viewpoint control from demonstrations.

\subsection{Learning Manipulation from Egocentric Demonstrations}
Several works augment robot policy learning by pretraining on large-scale egocentric human demonstrations, using the demonstrator's viewpoint to learn task-relevant visual representations and action structure, before grounding the model with a small amount of robot demonstrations.
Yang et al. proposed EgoVLA, which pretrains a vision-language-action model on egocentric human videos and then fine-tunes it with a small amount of robot demonstrations to obtain a robot policy~\cite{yangEgoVLALearningVisionLanguageAction2025a}.
Kareer et al. presented EgoMimic, a full-stack framework that co-trains a unified policy on egocentric human data and robot data to scale IL across diverse manipulation tasks~\cite{kareerEgoMimicScalingImitation2024b}.
% These approaches primarily leverage egocentric data for representation learning and/or policy scaling, and typically do not explicitly learn a visibility-driven viewpoint-repositioning policy to obtain target visibility under partial occlusions at deployment, which is the focus of our work.

Another line of work uses egocentric human demonstrations not as the primary supervision, but as complementary data during training, paired with robot demonstrations to broaden coverage.
Wang et al. proposed MimicPlay, which leverages large-scale human play videos to learn hierarchical plans and then trains low-level visuomotor control with a small amount of robot demonstrations for long-horizon manipulation~\cite{wangMimicPlayLongHorizonImitation2023a}.
Lepert et al. proposed Masquerade, which edits in-the-wild egocentric human videos into robotized demonstrations and uses them to pretrain visual representations and co-train policies, before fine-tuning with a small amount of robot demonstrations for robust long-horizon manipulation~\cite{lepertMasqueradeDataEditing2026a}.
% While these methods improve robustness and coverage, they generally assume that task-relevant visual cues are available from a fixed observation setup and do not treat viewpoint control as an action variable for resolving occlusions; in contrast, EgoReveal learns viewpoint-repositioning jointly with manipulation.

Human-only learning from egocentric demonstrations has also been explored, training policies purely from smart-glasses first-person human data and deploying them on real robots without requiring robot demonstrations.
Liu et al. proposed EgoZero, a minimal system that learns manipulation policies from Project Aria smart-glasses demonstrations with zero robot data~\cite{liuEgoZeroRobotLearning2025a}.
Gavryushin et al. presented MAPLE, which learns dexterous manipulation priors from egocentric videos by predicting contact points and detailed hand poses and transfers these representations to downstream manipulation policies~\cite{gavryushinMAPLEEncodingDexterous2025a}.
Guzey et al. presented AINA, which learns multi-fingered manipulation policies from first-person videos captured by smart glasses in everyday settings, without using any robot data~\cite{guzeyDexteritySmartLenses2025a}.
% Haldar and Pinto presented Point Policy, which enables training without teleoperation data by representing both observations and actions with semantically meaningful keypoints, and learns robot manipulation policies from offline human demonstration videos~\cite{haldarPointPolicyUnifying2025b}.
% Lepert et al. proposed Phantom, a framework that trains robot policies using only human videos and deploys them on real robots without robot demonstrations~\cite{lepertPhantomTrainingRobots2025a}.

These egocentric-demonstration-based manipulation methods typically do not explicitly learn viewpoint control policies for handling partial occlusions, which is a key distinction from EgoReveal.

\subsection{Learning Viewpoint Control from Demonstrations}
Recent works learn visibility-aware viewpoint control from human demonstrations by coupling viewpoint control with manipulation to handle occlusions and limited visibility, collecting such demonstrations via on-robot teleoperation using a HMD.
Chuang et al. presented AV-ALOHA, an HMD-based teleoperation system that learns an active vision policy jointly with manipulation by mapping the operator’s head motion to a dedicated camera-arm to maintain task-relevant visibility~\cite{chuangActiveVisionMight2025b}.
Xiong et al. proposed ViA, which learns task-relevant perceptual strategies such as search, tracking, and focusing from human demonstrations collected with an HMD using an actuated camera neck~\cite{xiongVisionActionLearning2025a}.
These HMD-based teleoperation approaches collect demonstrations by operating the robot and explicitly teleoperating the camera viewpoint through an actuated viewpoint mechanism such as a dedicated camera-arm or an actuated camera neck.
Such robot-operated, HMD-based data collection is costly and labor-intensive, whereas EgoReveal learns viewpoint motions from passively collected, natural egocentric demonstrations without on-robot teleoperation.

Robot-free human demonstration frameworks have also begun to incorporate visibility-aware viewpoint control.
Zeng et al. presented ActiveUMI, a portable HMD-based teleoperation kit that records an operator's head motion as supervision for viewpoint control while collecting end-effector commands using tracked, instrumented hand controllers that capture six-degree-of-freedom (6DoF) pose and discrete inputs~\cite{zengActiveUMIRoboticManipulation2025a}.
Yu et al. proposed EgoMI, which trains a memory-augmented policy to handle large viewpoint changes from synchronized head and hand trajectories captured using a HMD and tracked hand controllers~\cite{yuEgoMILearningActive2025a}.
These works still rely on dedicated capture devices and specialized data-collection interfaces to record head/hand coordination and retarget it to robots, whereas EgoReveal learns viewpoint motions directly from lightweight smart-glasses recordings of natural human demonstrations without requiring such dedicated interfaces.

\section{Preliminaries}

\subsection{Notation and Coordinate Frames}
Throughout this paper, we represent rigid-body transformations using homogeneous transformation matrices. 
The notation ${}^aT_b \in SE(3)$ denotes the rigid transformation from coordinate frame $b$ to coordinate frame $a$. In particular, ${}^aT_b$ transforms a point expressed in frame $b$ into frame $a$ as
\begin{equation}
\mathbf{x}^a = {}^aT_b \mathbf{x}^b,
\end{equation}
where $\mathbf{x}^a$ and $\mathbf{x}^b$ represent the homogeneous coordinates of the same 3D point in frames $a$ and $b$, respectively.

Each transformation matrix ${}^aT_b$ is written as
\begin{equation}
{}^aT_b =
\begin{bmatrix}
{}^aR_b & {}^a\mathbf{t}_b \\
\mathbf{0}^\top & 1
\end{bmatrix},
\end{equation}
where ${}^aR_b \in SO(3)$ denotes the rotation matrix and ${}^a\mathbf{t}_b \in \mathbb{R}^3$ denotes the translation vector from frame $b$ to frame $a$, expressed in frame $a$. Also, 
% The inverse transformation is given by
% \begin{equation}
% {}^bT_a = ({}^aT_b)^{-1},
% \end{equation}
chained transformations follow the standard composition rule
\begin{equation}
{}^aT_b \, {}^bT_c = {}^aT_c.
\end{equation}

% $T$ represents the pose in the world coordinate.

\subsection{Problem Formulation}
We focus on a tabletop pick task under partial occlusion, where the target object remains within the camera field of view but is only partially visible from the initial viewpoint.
Fully occluded or out-of-view targets are not addressed in this work and are left for future work.
% We focus on partial occlusions where the target remains within the camera field of view but is only partially visible at the initial viewpoint; handling fully-occluded or out-of-view targets is beyond the scope of this work.
% We consider a tabletop pick-and-place task where the target object is partially occluded at the initial viewpoint by a foreground object.
The goal is to complete a pick task reliably under partial occlusion by (i) viewpoint-repositioning to obtain target visibility and (ii) executing manipulation once the target becomes sufficiently visible.
Occlusions induce partial observability of task-relevant visual cues, tightly coupling perception and control.

The robot consists of a dedicated camera-arm equipped with an RGB-D camera on its end-effector and an independent manipulation-arm for object interaction. At each timestep $t$, the robot observes an RGB image $I_t$ and a depth image $D_t$ from the same camera viewpoint. Specifically, the RGB image $I_t$ serves a dual purpose: it acts as a conditioning signal for viewpoint planning and is used to estimate 2D object keypoints. Subsequently, the depth image $D_t$ is employed to unproject these 2D keypoints into 3D coordinates to facilitate manipulation.

% From the camera-arm's joint configuration $q_t^\mathrm{AV}$, the robot computes the camera pose $T_t^\mathrm{AV}$ via forward kinematics. The depth image $D_t$ is specifically utilized to estimate 3D keypoints of the target object for manipulation.

%The robot comprises a dedicated camera-arm with an end-effector-mounted RGB camera and a separate manipulation arm for object interaction.
%At each time step $t$, the robot observes an RGB image $I_t$ and computes the camera pose $T^\mathrm{cam}_t$ via forward kinematics from the camera-arm joint configuration $q^\mathrm{cam}_t$, and obtains a depth image $D_t$ when estimating 3D object keypoints for manipulation.

We assume access to markerless egocentric demonstrations captured by an egocentric camera (e.g., smart glasses).
Demonstrations are recorded without any handheld teaching devices, and the robot must execute the task without attaching markers to the manipulated objects.
A lightweight calibration step (e.g., using a fiducial marker) may be used to align the demonstration and robot coordinate frames.
During execution, the target object is specified by a reference image (and optionally a small set of annotated reference keypoints).
Demonstrations provide supervision for learning policies that operate from the robot-mounted camera during execution.

We learn two imitation policies from markerless egocentric demonstrations.
The Active Vision policy $\pi_\theta^\mathrm{AV}$ and the Manipulation policy $\pi_{\theta'}^\mathrm{MA}$ are coupled with a visibility-based switching module $g(\cdot)$ that decides whether to switch from active vision to manipulation at the current time step, where $\theta$ and $\theta'$ denote the learnable parameters.
Execution follows a two-phase receding-horizon closed-loop scheme.
% Starting from $t=0$, the robot executes $\pi_\theta^\mathrm{AV}$ and replans at each time step while $g(\cdot)$ indicates that the target is not yet sufficiently visible.
Starting from $t = 0$, the robot executes $\pi^\mathrm{AV}_\theta$ in a receding-horizon manner, where the policy generates a temporal chunk of actions at each inference. The robot follows this plan until the next replanning cycle, as long as $g(\cdot)$ indicates that the target is not yet sufficiently visible.
Once $g(\cdot)$ indicates that the target is sufficiently visible, the robot switches to $\pi_{\theta'}^\mathrm{MA}$ and executes it for the remainder of the episode without switching back, replanning in a receding-horizon manner after receiving new observations.

% After switching, we keep the camera-arm fixed at the resolved viewpoint.
% ; handling re-occlusion during manipulation is beyond the scope of this work.
% In each phase, the corresponding policy predicts a short action chunk from the current observation, and the robot executes only the next command before replanning.

\subsection{Action Chunking with Transformers (ACT)}
This subsection introduces Action Chunking with Transformers (ACT)~\cite{zhaoLearningFineGrainedBimanual2023a}, with which our Active Vision policy is implemented. ACT is an IL framework that predicts a sequence of future actions, known as an \emph{action chunk}, rather than a single action per time step. This approach helps mitigate compounding errors and enables smooth, fine-grained control. In the standard ACT formulation, the policy $\pi_\theta^{\mathrm{AV}}$ maps the current observation $o_t$ to a sequence of $H$ future actions $\hat{a}_{t:t+H-1}$:
% Action Chunking with Transformers (ACT)~\cite{zhaoLearningFineGrainedBimanual2023a} is an IL framework that predicts a sequence of future actions, known as an \emph{action chunk}, rather than a single action per time step. 
\begin{equation}
\hat{a}_{t:t+H-1} = \pi_\theta^\mathrm{AV} (o_t), 
\label{eq:act_1}
\end{equation}
where $\theta$ denotes the learnable parameters, $o_t$ represents the current observation, and $a_{t+1:t+H}$ is the predicted action chunk. To capture the multi-modality of demonstrations, ACT employs a Conditional VAE~\cite{sohnCVAE2015} to sample a latent style variable $z$. The policy, implemented using a ResNet18\cite{heResNet2016} backbone and a Transformer-based decoder, is optimized using a reconstruction loss between the predicted action chunk and the ground truth:
\begin{equation}
\mathcal{L}_\mathrm{recon} = | a_{t:t+H-1} - \hat{a}_{t:t+H-1} |_1, 
\label{eq:act_2}
\end{equation}
where $\hat{a}_{t:t+H-1}$ is the predicted action chunk.

\subsection{EgoZero}
This subsection introduces EgoZero~\cite{liuEgoZeroRobotLearning2025a} with which our Manipulation policy is implemented.
EgoZero is an IL framework that learns the Manipulation policy $\pi_{\theta'}^\mathrm{MA}$. 
The policy $\pi_{\theta'}^\mathrm{MA}$ is trained to predict a sequence of future end-effector actions $\hat{a} = a_{t:t+M-1}$ as 
\begin{equation}
\hat{a} = \pi_{\theta'}^\mathrm{MA} (s_t),
\label{eq:egozero_1}
\end{equation}
where $s_t=(O_0, a_{t-L+1:t})$ is the current state, $O_0 \in \mathbb{R}^{3 \times N}$ represents 3D keypoints on the manipulated object, $a_{t-L+1:t}$ is a history of past end-effector actions, each $a_i = [\mathbf{p}_i, q_i]^\top \in \mathbb{R}^4$ consists of a 3D position $\mathbf{p}_i$ and a binary gripper state $q_i$, $N$ represents the number of keypoints, $L$ is the number of past time steps considered, and $M$ denotes the number of predicted future time steps.

% where $s_t=(O_t, T_{t-L+1:t})$ is the current state, $O_t \in \mathbb{R}^{3 \times N}$ represents 3D keypoints on the manipulated object, $T_{t-L+1:t}$ is a history of past EEF poses, $N$ represents the number of keypoints, $L$ is the number of past time steps considered, and $M$ denotes the number of the predicted future time steps.

% \subsection{EgoZero}
% This subsection introduces EgoZero~\cite{} with which our manipulation policy is implemented. EgoZero is an imitation learning framework that learns manipulation policies. The policy is trained to predict a sequence of future end-effector (EEF) actions of the robot arm $a=A_{t+1:t+M}$ given the target object 3D keypoints $O_t \in \mathbb{R}^{3 \times N}$ and a history of past EEF states $A_{t-L+1:t}$ as state $s_t=(O_t, A_{t-L+1:t})$, where $N$ represents the number of keypoints, $L$ is the number of past time steps considered, and $M$ denotes the number of future time steps predicted:
% %
% \begin{equation}
% \pi_{\theta}^{\mathrm{MA}} : (O_t, A_{t-L+1:t}) \rightarrow A_{t+1:t+M}.
% \end{equation}
%
The policy is trained with a closed loop Transformer policy~\cite{vaswaniTransformer2017} using behavior cloning over our dataset $\mathcal{D}$. The policy predictions are modeled as the mean of a normal distribution and optimized by minimizing the negative log likelihood with a fixed variance parameter $\sigma$.
\begin{equation}
\theta' = \arg \min_{\theta'} 
\mathbb{E}_{s, a \sim \mathcal{D}}
\left[
\frac{\|\pi^{\mathrm{MA}}_{\theta'}(s) - a\|^2}{2\sigma^2}
\right].
\end{equation}

\section{Method}

\subsection{Overview}
\begin{figure}[t]
    \centering
    \includegraphics[width=1.0\linewidth]{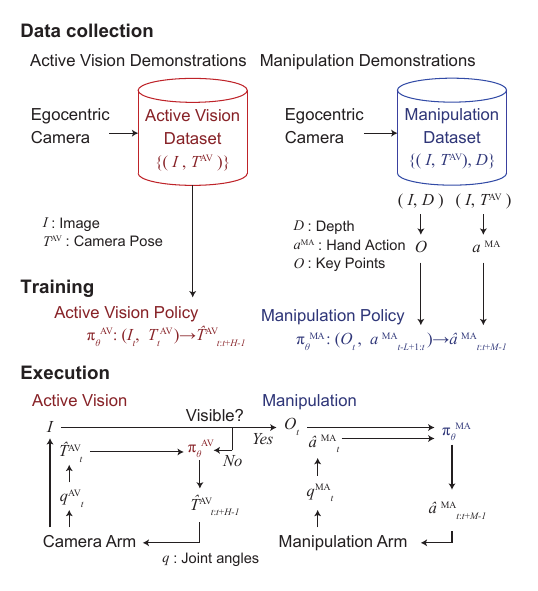}
    \caption{EgoReveal framework.}
    \label{fig:method}
\end{figure}

This paper proposes EgoReveal, a framework that learns viewpoint-repositioning and manipulation actions using egocentric human demonstration videos (Fig.~\ref{fig:method}). The goal is to enable a robot to reposition its viewpoint under occlusions for manipulating occluded objects. 

EgoReveal includes two imitation policies. The first is the Active Vision policy that predicts a camera-arm pose trajectory capable of avoiding occlusions, given the current visual observation and the current state of the camera-arm. The second is the Manipulation policy that predicts control commands from 3D keypoints of the object to be grasped, extracted from the camera view established at the end of the Active Vision policy.
These policies are trained to imitate human demonstrations collected through smart glasses, enabling learning from natural egocentric human demonstrations acquired without teleoperation or specialized teaching devices.

In addition, to determine when to switch between the two policies, namely, when occlusion has been sufficiently resolved, and the system can transition to object manipulation, EgoReveal introduces an Occlusion Resolution Assessment (ORA) module. This module evaluates whether the target object is adequately visible and decides the appropriate timing for switching from the Active Vision policy to the Manipulation policy.
% The following subsections describe the Active Vision policy and the Manipulation policy in detail in Secs. XX and XX, respectively. The Occlusion Resolution Assessment module is then described in Sec. XX.

\begin{comment}
% システム概要
Our proposed pipeline comprises a dual-component architecture consisting of an Active Vision module and a Manipulation module. The Active Vision module is designed to maneuver the robot to mitigate occlusions and ensure optimal observation of the target object, whereas the Manipulation module leverages pre-existing models to execute pick-and-place operations. Both modules are trained to imitate human demonstrations acquired via smart glasses; specifically, the active vision policy is deployed onto a dedicated camera-arm, while the manipulation policy is integrated into the manipulation arm. The remainder of this chapter is organized as follows: Section 3.1 describes the Project Aria smart glasses \cite{engelProjectAriaNew2023} utilized for data collection, and Sections 3.2 and 3.3 provide a detailed elaboration on the Active Vision and Manipulation modules, respectively.
\end{comment}

\begin{comment}
\subsection{Project Aria Smart Glasses}
The Project Aria smart glasses serve as an ideal platform for collecting first-person human demonstration data, housing visual sensors such as fisheye RGB, SLAM, and eye-tracking cameras, as well as non-visual sensors like IMUs and microphones, within a unified device. Furthermore, the Machine Perception Service (MPS) is employed to process observational data, enabling the accurate extraction of device pose, hand pose, and eye-gaze. For both modules, we utilize inputs from a single fisheye RGB camera and two SLAM cameras. In the Active Vision module, we derive camera intrinsics and extrinsics via MPS, ultimately compiling a dataset of RGB images and device poses, denoted as $(I_t^{raw}, T_t^{dev})$. Conversely, for Manipulation, we additionally extract palm pose information to construct a final record comprising RGB frames, device poses, and palm poses, represented as $(I_t^{raw}, T_t^{dev}, H_t^{palm})$.
\end{comment}

\subsection{Active Vision Policy}
We implement the Active Vision policy $\pi^{AV}_{\theta}$ by adapting the ACT framework, a widely adopted visuo-motor policy, to predict camera-arm trajectories.
% We implement the Active Vision policy $\pi_{\theta}^{\mathrm{AV}}$ by adapting the ACT framework to predict camera-arm trajectories.
While the network architecture maintains the core design of ACT, we modify the input and output spaces to accommodate the specific requirements of our active vision task. Specifically, while the original ACT implementation typically processes visual feedback from multiple cameras to output joint actions, our policy utilizes a single-view setup. Following the formulation in Eq. \ref{eq:act_1}, we define the observation as $o_t^{\mathrm{AV}} = (I_t, T_t^\mathrm{AV})$, where $I_t$ is the current rectified RGB image from the camera-arm and $T_t^\mathrm{AV}$ is the current camera-arm pose. The policy then outputs an action chunk $a_t^{\mathrm{AV}} = T_{t:t+H-1}^\mathrm{AV}$, which represents a sequence of future camera poses instead of manipulator joint states.
To efficiently acquire human demonstrations, we collect first-person-view RGB video frame sequences captured by the fisheye camera embedded in the wearable glasses. To remove lens distortion and provide standardized inputs for policy learning, the fisheye images are rectified by reprojecting them onto a virtual pinhole camera model, producing rectified observations denoted as $I_t$. The corresponding camera-arm poses $T_t$ are obtained offline from the smart glasses during the data collection process.

\subsection{Occlusion Resolution Assessment Module}
To switch the processing from the Active Vision policy to the Manipulation policy, we propose an Occlusion Resolution Assessment Module that determines whether the observation camera has sufficiently resolved the occlusion and can clearly observe the target object for grasping.

Given the current image observation $I_t$, 
the proposed method first applies OwlViT~\cite{mindererSimpleOpenVocabulary2022} to perform object detection, given a target reference image. To robustly detect the reference object, we perform filtering based on two criteria: (1) the similarity score computed by OwlViT's image-derived embeddings for coarse filtering and (2) the matching score in HSV color space for fine filtering. 

For the first filtering, let $b_i$ denote the $i$-th bounding box candidate with width $w_i$, height $h_i$, and matching score $s_i$. Let $\tau_w$ and $\tau_h$ be predefined thresholds for width and height. Among the candidates that satisfy the size constraints, we select the optimal bounding box $b^*$ as
\begin{equation}
b^{*}
=
\arg\max_{b_i \in \mathcal{B}}
\left\{
s_i
\;\middle|\;
w_i \le \tau_w,\;
h_i \le \tau_h
\right\}.
\end{equation}

For the second filtering, SAM3~\cite{carion2025sam3segmentconcepts} is employed to segment the selected bounding box and extract the target object region.
% segmentation is applied to the selected bounding box to extract the target object region. 
For this region, we compute a matching score $s_1$ by measuring the Bhattacharyya distance between the two-dimensional histograms of the reference image and the current region in the HS channels of the HSV color space. The subsequent processing proceeds only if $s_1$ exceeds a threshold $\tau_{s0}$. 

To determine whether occlusion has been sufficiently resolved, we compute the masked area $S_t$ of the segmented target object region. 
While the Active Vision policy actively adjusts the camera pose, we monitor the temporal trend of the observed masked area over the most recent $k$ frames. 
Specifically, we compute the slope between time steps $t-k$ and $t$ as $\Delta S^{(k)} = (S_t - S_{t-k})/k.$
When $\left| \Delta S^{(k)} \right| < \tau_{s1}$, indicating that the observed area has converged and no longer exhibits significant change, we regard the occlusion as sufficiently resolved and switch from the Active Vision policy to the Manipulation policy.

\subsection{Manipulation Policy}
% Our Manipulation policy extends the EgoZero framework. 
% Unlike the original EgoZero, which assumes a stationary egocentric camera as input, EgoReveal utilizes a camera mounted on the robot arm and performs active viewpoint adaptation to get target object 3D keypoints $O_{t=0}$. 
% This setting introduces a more challenging problem, as the visual observations vary dynamically with the motion of the camera.
% We describe how we process and the corresponding coordinate transformations required under these challenging settings as follows.

We adopt EgoZero as our Manipulation policy because it enables learning robotic tasks exclusively from first-person human demonstration data.
% We adopt EgoZero as our Manipulation policy, a framework that learns robotic tasks exclusively from human demonstration data.

% The Manipulation policy is trained to predict a sequence of future end-effector (EEF) actions of the robot arm $A_{t+1:t+M}$ given the target object 3D keypoints $O_t \in \mathbb{R}^{3 \times N}$ and a history of past EEF states $A_{t-L+1:t}$, where $N$ represents the number of keypoints, $L$ is the number of past time steps considered, and $M$ denotes the number of future time steps predicted, respectively. 
% %
% \begin{equation}
% \pi_{\theta}^{\mathrm{MA}} : (O_t, A_{t-L+1:t}) \rightarrow A_{t+1:t+M}.
% \end{equation}
% %
% Our Manipulation policy extends the EgoZero framework~\cite{liuEgoZeroRobotLearning2025a}. Unlike the original EgoZero, which assumes a fixed third person camera as input, our approach utilizes a camera mounted on the robot arm and performs active viewpoint adaptation. This setting introduces a more challenging problem, as the visual observations vary dynamically with the motion of the camera.

% \textbf{Visual Observation and Processing} 
To estimate the position of the target object for manipulation, the system starts from the current RGB observation $I_0$ immediately after switching policies using ORA. 
GroundingDINO~\cite{liuGroundingDINO2024} is first applied to $I_0$ to perform object detection and obtain the bounding box of the target object. 
Based on the detected bounding box, we crop $I_0$ to obtain the cropped image $I_0^\mathrm{crop}$.
Subsequently, the DIFT model~\cite{tangEmergentCorrespondenceDiffusion2023} identifies 2D keypoints within the cropped image by referencing an annotated template.
These points are then unprojected into 3D space using camera intrinsics and depth information, resulting in the object 3D keypoints $O_0$.
The derived keypoints $O_0$ are combined with the historical sequence of robot states $T_{t-L+1:t}$, constitute the input $s_t$ to the policy. 
Finally, the manipulation actions are predicted according to the formulation in Eq. \ref{eq:egozero_1}.
% the future EEF poses as 
% \begin{equation}
%     a_t^\mathrm{MA} = \pi_t^\mathrm{MA}(s_t) = T_{t+1:t+M}^\mathrm{MA}
% \end{equation}

\begin{comment}
\textbf{Policy Training}
The Manipulation policy takes the object 3D keypoints $O_t$ and the historical sequence of robot states $A_{t-L+1:t}$ as the state $s_t=(O_t, A_{t-L+1:t})$ as input. Then the policy is trained to output $A_{t+1:t+M}$ as the corresponding action $a$. Following EgoZero~\cite{}, we train a closed loop Transformer policy~\cite{} using behavior cloning over our dataset $\mathcal{D}$. The policy predictions are modeled as the mean of a normal distribution and optimized by minimizing the negative log likelihood with a fixed variance parameter $\sigma = 0.1$~\cite{}.
%
\begin{equation}
\theta = \arg \min_{\theta} 
\mathbb{E}_{s, a \sim \mathcal{D}}
\left[
\frac{\|\pi^{\mathrm{MA}}_\theta(s) - a\|^2}{2\sigma^2}
\right]
\end{equation}
%
\end{comment}

\subsection{Coordinate Transformation}
Unlike the original ACT, EgoReveal involves multiple coordinate systems that must be consistently aligned for data collection, policy training, and inference. 
% Specifically, the inputs and outputs handled by the policy described above are defined in the camera coordinate system used during data acquisition. Therefore, they must be transformed into the end effector coordinate system of the robot arm for execution.
Specifically, the inputs and outputs handled by the policies described above are defined using the camera-arm pose expressed in the robot base coordinate system.
However, since the physical robot is not present during human demonstrations, it is not straightforward to share a common coordinate system between the training data collection phase that uses egocentric video, and the training/inference phase, which involves the real robot. To address this issue, we introduce a virtual base coordinate system defined by a visible marker and denote the camera pose in this frame as ${}^mT_c$, which represents the camera pose within the marker coordinate system. By introducing the marker coordinate system, the relationship between the camera coordinate system and the robot base coordinate system can be expressed as
$${}^mT_c = {}^mT_w \cdot {}^wT_c,$$
where ${}^wT_c$ denotes the continuous camera trajectory provided by the device’s SLAM system relative to a fixed world frame $w$, and ${}^mT_w$ represents the static transformation from the world frame to the marker frame.

% Unlike the original EgoZero, EgoReveal involves multiple coordinate systems that must be consistently aligned throughout data collection, policy training, and inference. 
For manipulation, whereas the original EgoZero relies on observations from a fixed camera, our method processes dynamically moving egocentric camera observations and employs an additional robot arm to actively avoid occlusions.
Specifically, the Manipulation policy produces outputs in the same camera coordinate system as the Active Vision policy. However, the actual target object grasping is performed by a separate manipulation-arm. Therefore, the predicted camera poses must be transformed into the robot base coordinate of the manipulation-arm.
We achieve this by applying a chain of coordinate transformations as follows. 
\begin{equation}
{}^{\mathrm{ma}}T_{\mathrm{cam}}
=
{}^{\mathrm{ma}}T_{\mathrm{av}}
\cdot
{}^{\mathrm{av}}T_{\mathrm{cam\_eef}}
\cdot
{}^{\mathrm{cam\_eef}}T_{\mathrm{cam}},
\end{equation}
Here, ${}^{\mathrm{cam\_eef}}T_{\mathrm{cam}}$ represents the extrinsic calibration between the camera and the end-effector of the camera-arm. The transformation from camera-arm end-effector to base active vision arm ${}^{\mathrm{av}}T_{\mathrm{cam\_eef}}$ is obtained from the forward kinematics of the active vision arm, and ${}^{\mathrm{ma}}T_{\mathrm{av}}$ represents the calibration between the base frames of the two robot arms.

\subsection{Execution Pipeline}
At runtime, we execute a receding-horizon closed-loop scheme with two learned imitation policies, $\pi_\theta^{\mathrm{AV}}$ and $\pi_{\theta'}^{\mathrm{MA}}$, coupled by a visibility-based switching module.
Execution starts in the active-vision mode: the robot first applies $\pi^{\mathrm{AV}}$ to reposition the camera viewpoint and regain line of sight to the target under occlusion.
In the active-vision mode, the policy takes the current rectified view $I_t^\mathrm{AV}$ and the current camera pose $T_t^\mathrm{AV}$ (mounted on the camera-arm end-effector) as input, and predicts a short-horizon chunk of target camera poses $T_{t+1:t+H}^\mathrm{AV}$.
The camera-arm robot follows the predicted action chunk sequentially. A replanning cycle is initiated when the number of remaining steps in the current chunk falls below a predefined threshold. At each replanning step, the policy generates a new trajectory based on the updated observation, and the target poses are converted to camera-arm joint targets via IK for execution.

Given the current rectified observation $I_t^\mathrm{AV}$, the switching module $g(\cdot)$ (implemented by ORA) evaluates target visibility and outputs a binary decision indicating whether the target is sufficiently visible.
When ORA outputs that the target is sufficiently visible, the system switches to the manipulation mode and applies $\pi_{\theta'}^{\mathrm{MA}}$ to perform object interaction.
This switch is performed at most once per episode: after switching to $\pi_{\theta'}^{\mathrm{MA}}$, the system does not switch back to $\pi_\theta^{\mathrm{AV}}$.
After switching, we hold the camera-arm at the last active-vision pose to maintain a stable viewpoint; handling re-occlusion during manipulation is beyond the scope of this work.

In the manipulation mode, immediately after switching from active vision, we compute object 3D keypoints $O$ once from the first camera-arm observation after the switch, $I_t^\mathrm{AV}$ (and depth $D_t$ when available).
We detect the target with GroundingDINO, match 2D keypoints using DIFT, and unproject them to 3D using the camera intrinsics and depth.
We keep these keypoints fixed during the remainder of the episode and apply $\pi_{\theta'}^{\mathrm{MA}}(O, A_{t-L+1:t})$ to predict an end-effector action chunk; the next action is transformed into the manipulation-arm base frame and executed via IK until task completion, with replanning at each step using the updated robot state history $A_{t-L+1:t}$.

%%%%%%%%%%%%%%%%%%%%%%%%%%%%%%%%%%%%%%%%%%%%%%%%%%%%%%%%%%%%%%%%%%%%%%%%%%

\section{Experimental Settings}

\subsection{Scenarios and Tasks} 

\begin{figure}[t]
    \vspace*{8pt}
    \centering
    \includegraphics[width=1.0\linewidth]{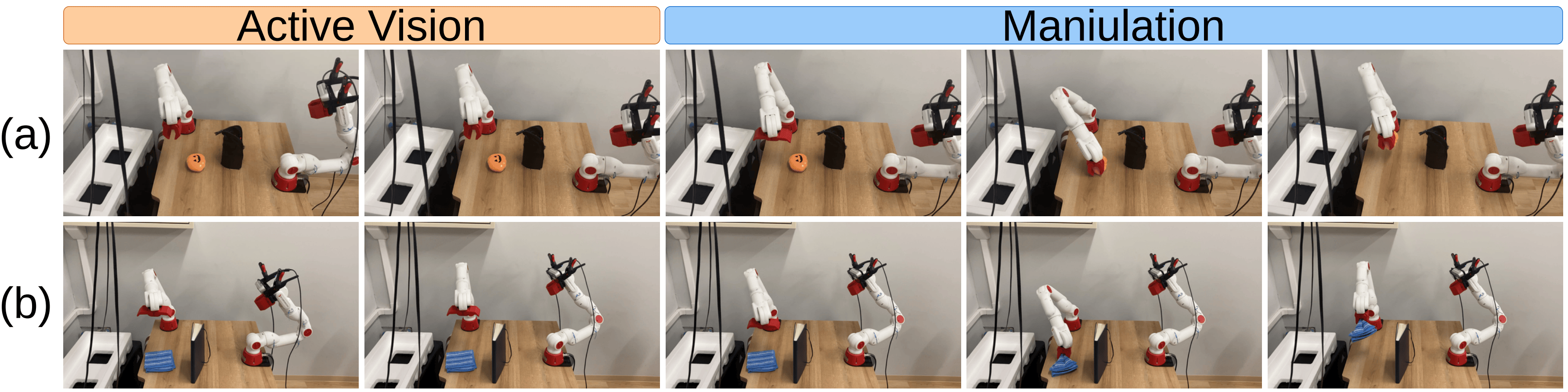}
    \caption{Experimental scenarios. (a) SoftToy, (b) Towel.}
    \label{fig:inference}
\end{figure}

\begin{figure}[t]
    \vspace*{8pt}
    \
    \centering
    \includegraphics[width=1.0\linewidth]{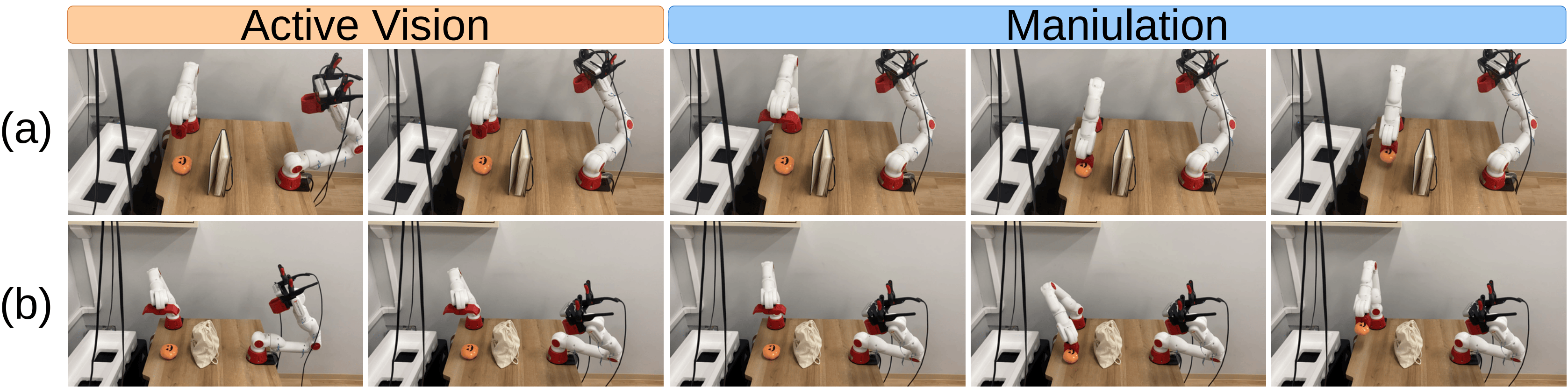}
    \caption{Experimental scenarios with unseen occlusions (a) SoftToy partially occluded by a book, (b) SoftToy partially occluded by a bag.}
    \label{fig:inference_unified}
\end{figure}

\begin{figure}[t]
    \centering
    \includegraphics[width=0.95\linewidth]{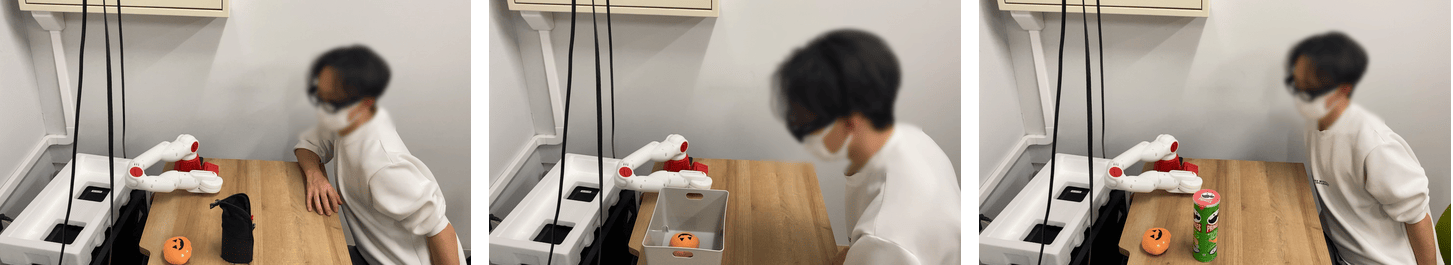}
    \caption{Data Collection for AV against Diverse Occlusions.}
    \label{fig:data_collection_unified}
\end{figure}

To evaluate the proposed method in performing object manipulation while resolving occlusions, we design a pick task with occluded target objects to be grasped. 
%To evaluate robustness across diverse occluders and grasping targets
We prepare the following two experimental scenarios, as illustrated in Fig.~\ref{fig:inference}. 
%
% \textbf{Soft Toy}: A soft toy partially occluded by a pencil case is manipulated after the camera moves laterally to resolve the occlusion, \textbf{Towel}: A towel partially occluded by a book is manipulated after the camera moves laterally to resolve the occlusion.
\textbf{Soft Toy}: A soft toy partially occluded by a pencil case, and \textbf{Towel}: A towel partially occluded by a book. Both are manipulated after the camera moves laterally to resolve the occlusion.
Furthermore, to evaluate the generalization capability to unseen occlusions, we trained a unified policy using data collected through active perception across three distinct occluders (Fig. \ref{fig:data_collection_unified}). This policy was then tested on two novel occluders (Fig. \ref{fig:inference_unified}) to verify its robustness against previously unseen occlusion scenarios.

In all experiments, we focus on partial occlusion, ensuring that the target object is only partially visible from the initial observation point. To reproduce this condition, the camera-arm is initialized at a pose that produces an occluded view, while the manipulation-arm is initialized at a random horizontal position approximately 20 cm above the target object. Furthermore, in each episode, the relative positions of the occluder and the target object on the table are randomized.

We define success for each phase of the task. The active vision phase is considered successful if the target object is entirely unoccluded at the moment $g(\cdot)$ signals the switch to manipulation. The overall task is successful if the manipulation-arm subsequently achieves a stable grasp and lifts the target object.

% \begin{figure}[htbp]
%     \centering
%     \includegraphics[width=0.8\linewidth]{fig/hardware.JPEG}
%     \caption{Hardware Setup}
%     \label{fig:sample}
% \end{figure}

\subsection{Hardware}
We utilize two robot arms (CRANE-X7, RT Corporation), each equipped with a 7DoF arm and a gripper. One robot is dedicated to the manipulation task, while the other functions as a camera mount for observation. The two robots are arranged with a 90-degree angular offset relative to each other.
Regarding the sensory setup, we employ Project Aria~\cite{somasundaramProjectAria2023}, smart glasses for the Active Vision policy and Occlusion Resolution Assessment Module. Since the Active Vision policy relies on RGB input, we use Project Aria to ensure consistency between data collection and inference. In addition, as the Manipulation policy requires depth observation, we also incorporate an Intel RealSense D435 camera.

\subsection{Data Collection}

\begin{figure}[t]
    \vspace*{8pt}
    \centering
    \includegraphics[width=0.95\linewidth]{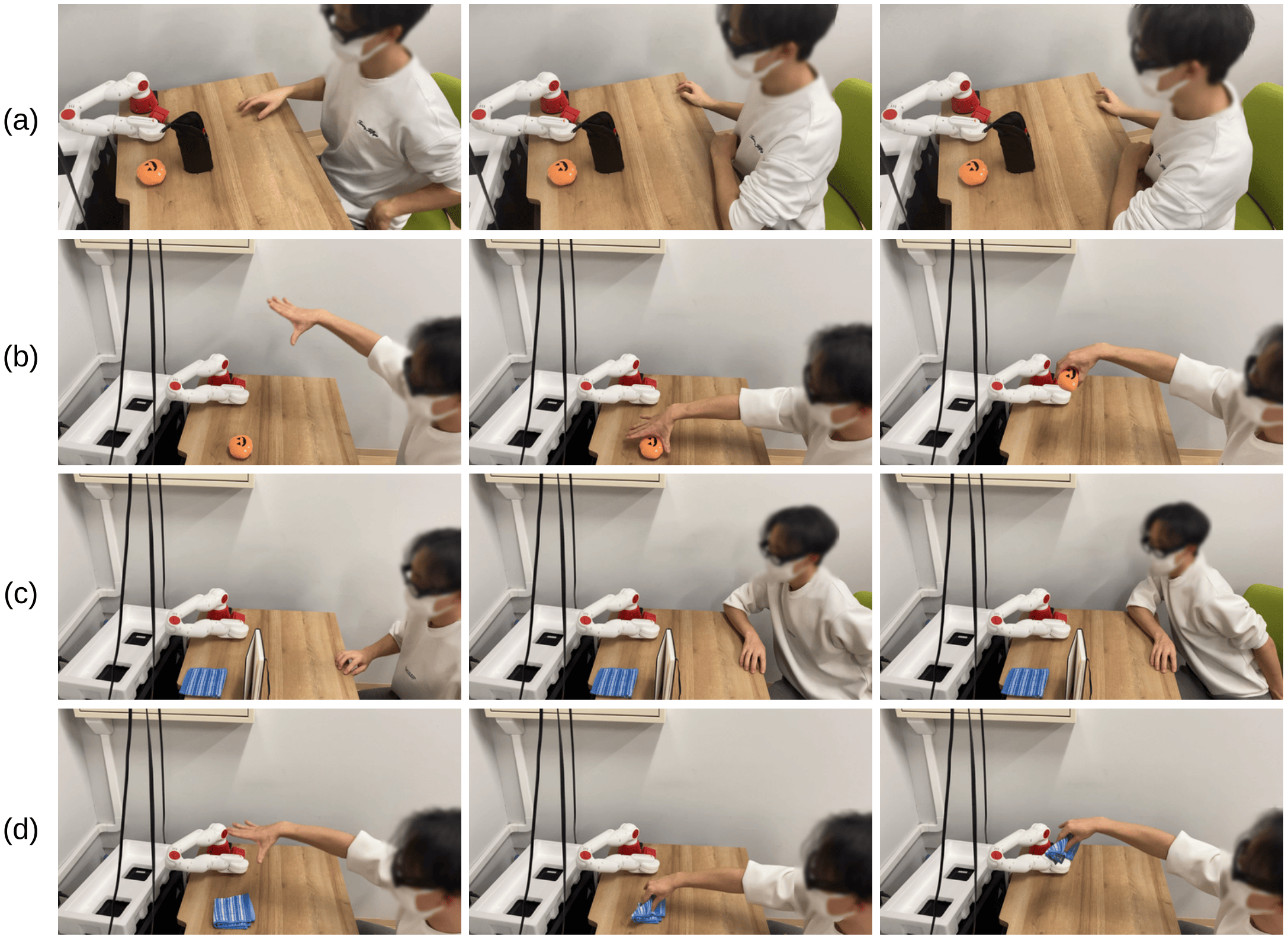}
    \caption{Data Collection. (a) SoftToy AV, (b) SoftToy MA, (c) Towel AV, (d) Towel MA.}
    \label{fig:data_collection}
\end{figure}

For both the Active Vision task and the Manipulation task, we recorded first-person view demonstrations in which a human operator performed the required operations (Fig.~\ref{fig:data_collection}).
In the Active Vision task, we collected 100 episodes where a human operator wearing smart glasses performed a ``move-around'' motion to resolve partial occlusion. Throughout these recordings, we randomized the relative positions of the objects to create diverse spatial configurations and diversified the operator's head trajectories to ensure a wide range of viewpoints and motion patterns.
In the Manipulation task, we gathered 100 episodes of grasping and lifting the target object. These demonstrations were performed without any occluders present. During this process, both the position of the target object and the pose of the operator's head were randomized.

\subsection{Baseline Methods}
This paper addresses a novel task in which a robot learns from human demonstrations to resolve partial occlusions and manipulate a target object. To the best of our knowledge, there is no existing method that directly tackles this combined problem.
Therefore, we adopt EgoZero, a method that learns the Manipulation policies from human demonstrations, as our baseline approach.

\begin{comment}
Given the lack of existing studies that derive both active vision and manipulation policies from egocentric human demonstrations, we employ EgoZero as our primary baseline. In this context, EgoZero serves as an ablated version of our framework, excluding the active vision component. By comparing these approaches, we investigate whether the introduction of active vision enables the system to effectively circumvent occlusions and acquire the visual information necessary for successful manipulation.    
\end{comment}

\subsection{Implementation Details}
% We trained both policies with a single NVIDIA RTX 6000 Ada Generation GPU with a batch size of 8 and 10 epochs. We used Adam~\cite{kingmaAdam2015} Optimizer with a learning rate of $1 \cdot 10^{-5}$. 
% % We set $\lambda_{\alpha}=0.7, \lambda_{\beta}=0.3$.
Both the Active Vision and Manipulation policies were trained using a single NVIDIA RTX 6000 Ada Generation GPU with a batch size of 8. For the Active Vision policy, we employed the AdamW~\cite{loshchilov2018decoupled} optimizer with a learning rate of $1 \cdot 10^{-5}$ for 100,000 steps. For the Manipulation policy (EgoZero), we utilized the same optimizer with a learning rate of $1 \cdot 10^{-4}$ for 300,000 steps.

%%%%%%%%%%%%%%%%%%%%%%%%%%%%%%%%%%%%%%%%%%%%%%%%%%%%%%%%%%%%%%%%%%%%%%%%%%%%

\section{Results}

We conducted the following experiments.
(1) A comparison with the baseline method EgoZero to evaluate the effectiveness of the proposed framework.
(2) An ablation study to investigate the contributions of Active Vision module and ORA module.

\subsection{Comparison with the Baseline Method}

% \begin{table}[t] 
%     \centering
%     \caption{Comparison with the baseline method and ablation study.}
%     % \caption{Comparison with the baseline method and ablation study on success rate (SR) and task completion time (TCT). Note that TCT for EgoZero was not recorded since the method failed to complete the manipulation task in all trials.}
%     \label{tab:sr_comparison}
%     \begin{tabular}{lcc}
%         \toprule
%         \multirow{2}{*}{Method} & Soft Toy & Towel\\
%         \cmidrule(l{2pt}r{3pt}){2-2} \cmidrule(l{2pt}r{3pt}){3-3} 
%           & Success Rate $(\%)$ & Success Rate $(\%)$ \\
%         % Method & Soft Toy $(\%)$ & Towel $(\%)$\\
%         \midrule
%         \textbf{Ours} & \textbf{85} & \textbf{90} \\
%         EgoZero (Ours w/o AV)~\cite{liuEgoZeroRobotLearning2025a} & 35 & 45 \\
%         Ours w/o ORA & 55 & 70 \\
%         \bottomrule
%     \end{tabular}
% \end{table}

\begin{table}[t] 
    \vspace*{8pt}
    \centering
    \caption{Comparison with the baseline method and ablation study. \\ SR denotes success rate.}
    \label{tab:sr_comparison}
    \small % 表全体の文字サイズを少し小さくすると収まりやすくなります
    \begin{tabular}{lccc} 
        \toprule
        \multirow{3}{*}{Method} & Soft Toy & Towel & \makecell{Unseen \\ Occluders} \\ % 見出しも改行
        \cmidrule(lr){2-2} \cmidrule(lr){3-3} \cmidrule(lr){4-4} 
        & SR (\%) & SR (\%) & SR (\%) \\
          % & \makecell{Success \\ Rate (\%)} & \makecell{Success \\ Rate (\%)} & \makecell{Success \\ Rate (\%)} \\ % ここで改行
        \midrule
        \textbf{Ours} & \textbf{85} & \textbf{90} & \textbf{70} \\ 
        EgoZero (Ours w/o AV)~\cite{liuEgoZeroRobotLearning2025a} & 35 & 45 & 30 \\ 
        Ours w/o ORA & 55 & 70 & 50 \\ 
        \bottomrule
    \end{tabular}
\end{table}

To evaluate the effectiveness of the proposed method in object manipulation under partial occlusion, we conducted a comparison with EgoZero.
% A trial was considered successful if the target object was successfully grasped and lifted.

\begin{figure}[t]
    \centering
    \includegraphics[width=0.95\linewidth]{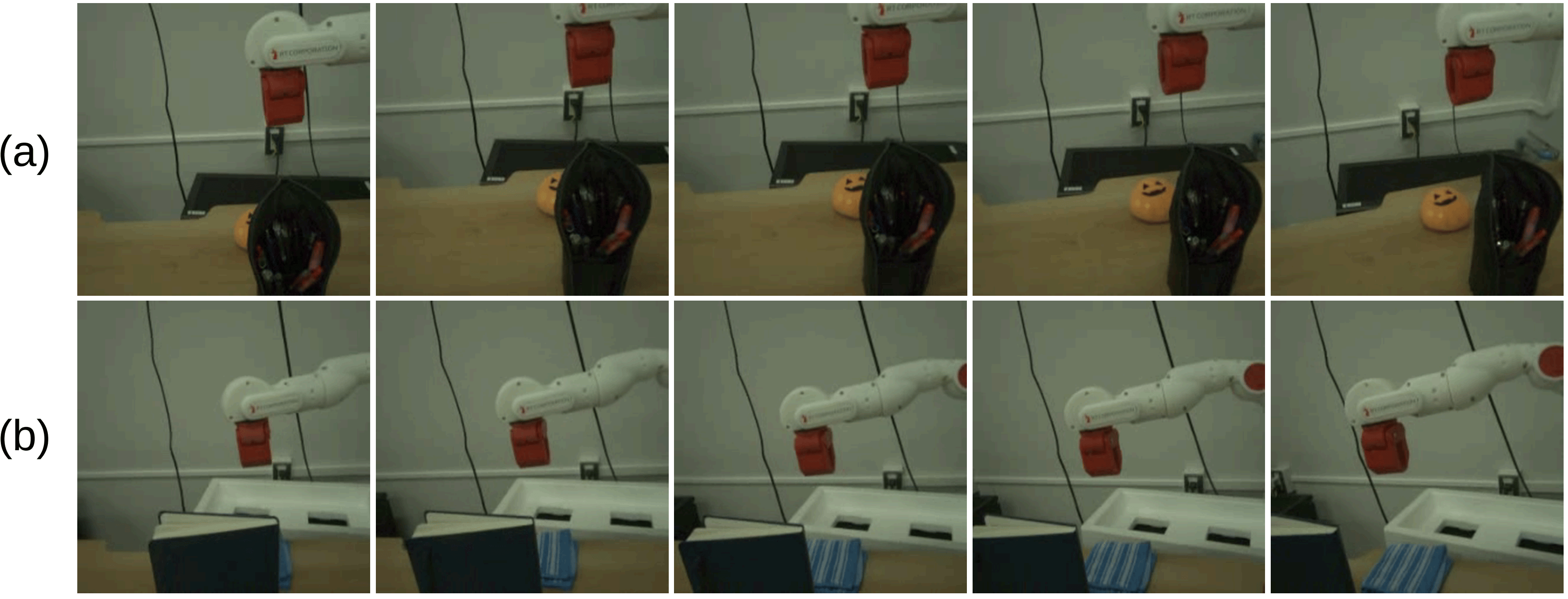}
    \caption{Sequence of Observations at Inference Time. (a) SoftToy, (b) Towel.}
    \label{fig:inference_obs}
\end{figure}

\begin{figure}[t]
    \vspace*{8pt}
    \centering
    \includegraphics[width=0.6\linewidth]{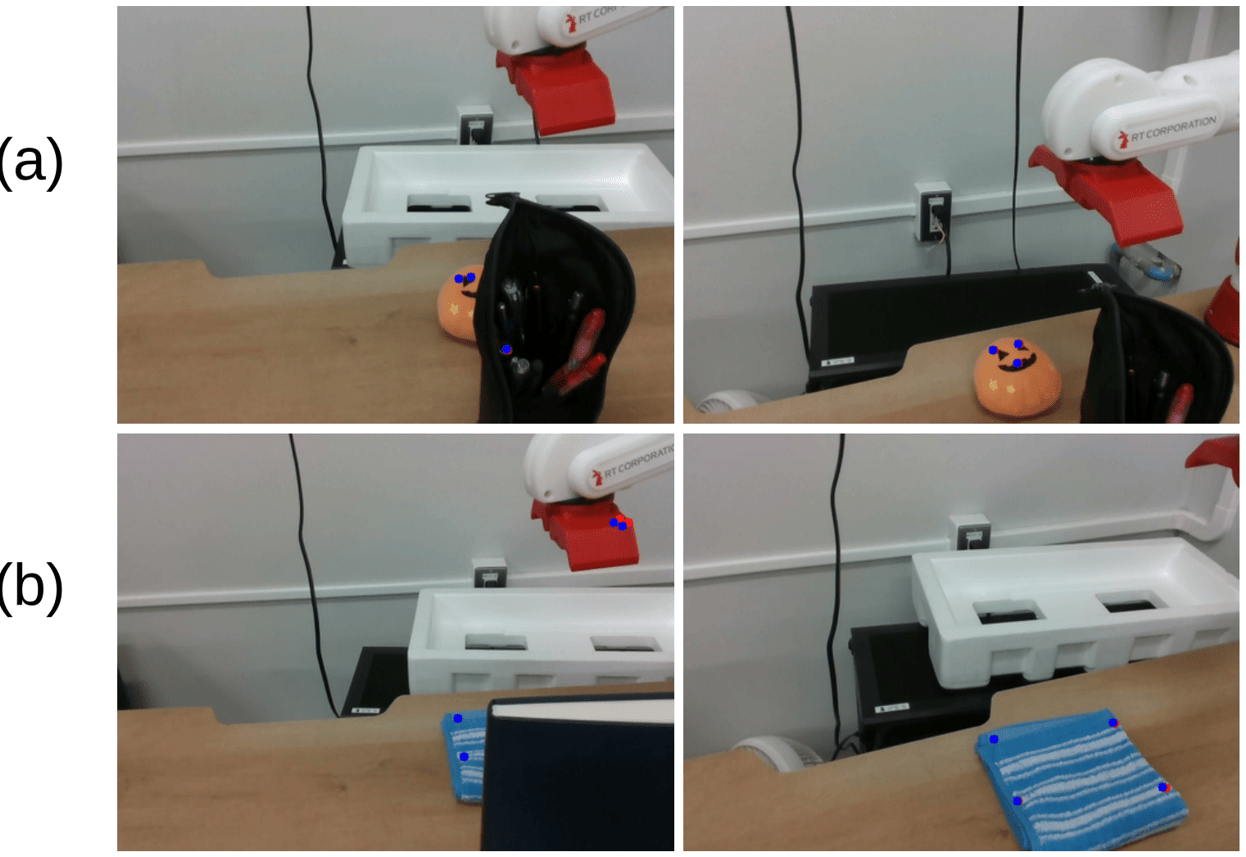}
    \caption{Keypoints Extraction for Manipulation. (a) SoftToy, (b) Towel. Left: Ours w/o AV; Right: Ours.}
    \label{fig:keypoint}
\end{figure}

As shown in Table~\ref{tab:sr_comparison}, the proposed method achieved higher success rates compared to the baseline. These results suggest that the introduced components effectively improve performance in occluded environments.
% Fig.~\ref{fig:inference_obs}, Fig.~\ref{fig:keypoint} illustrates the observation and processing pipeline of both the proposed method and EgoZero
Fig.~\ref{fig:inference_obs} illustrates the transition of observations during the Active Vision phase of our proposed policy. Meanwhile, Fig.~\ref{fig:keypoint} compares the keypoint extraction results of our method after occlusion avoidance against those of the baseline, EgoZero.
As shown in these figures, our method, which incorporates the Active Vision policy, effectively avoids occlusions and maintains the target object within the field of view. As a result, it enables accurate extraction of object keypoints required for manipulation.
In contrast, EgoZero is constrained by a fixed camera viewpoint and lacks the capability to avoid occlusion. Consequently, it fails to obtain reliable feature representations and frequently assigns keypoints to occluding objects. The baseline often attempts grasping actions at locations unrelated to the actual target, leading to task failure.

\subsection{Ablation Study}

To evaluate the contributions of the Active Vision module and the Occlusion Resolution Assessment Module, we compared the full proposed method ({\bf{Ours}}) with the following two variants.
{\bf{Ours w/o AV}}, a method without the Active Vision policy (same as original EgoZero), and 
{\bf{Ours w/o ORA}}, our method without the Occlusion Resolution Assessment Module, where switching between the Active Vision policy and the Manipulation policy is performed after a fixed time interval $T$. This interval $T$ was set to the average time at which the ORA module issued a termination signal during successful trials of our full method (Ours). We evaluated the performance of all methods using the success rate.

\begin{comment}
The success criteria were defined differently for each comparison. In the comparison with Ours w/o AV, a trial was considered successful if the target object was successfully grasped and lifted. In the comparison with Ours w/o ORA, a trial was deemed successful if the policy switch occurred when the target object was completely unoccluded.    
\end{comment}

%In the case of Ours w/o AV, the system failed to complete the manipulation task in all trials. We compared each variant with our method (Ours) based on the success rate.

As shown in Table~\ref{tab:sr_comparison}, the full proposed method (Ours), which includes both Active Vision and the Occlusion Resolution Assessment Module, achieved the highest success rate. These results demonstrate the effectiveness of both components in enabling robust manipulation under occlusion.

\subsection{Evaluation on Unseen Occlusion Scenarios}
To evaluate generalization to unseen occlusions, we conducted experiments with occluding objects not present during training and measured the success rate of the proposed method.
As shown in Table~\ref{tab:sr_comparison}, our method achieved higher success rate compared with baseline method and our method without Occlusion Resolution Assessment Module. These results demonstrate that our method robustly avoids unseen occlusions while successfully acquiring the information necessary for target-object manipulation. Furthermore, these findings indicate that Active Vision policy effectively discerns the spatial relationship between unknown objects and the target via visual feedback to infer optimal movement directions and poses.

%%%%%%%%%%%%%%%%%%%%%%%%%%%%%%%%%%%%%%%%%%%%%%%%%

\section{Discussion}
Although our experiments focused on a fixed-base manipulator with a gripper under partial occlusions, EgoReveal is inherently modular and can be extended to more general scenarios.

\begin{comment}

{\bf{Applicability to Mobile Robots}.} In hardware setups where the robot base itself moves, such as mobile manipulators, our framework can be applied by incorporating appropriate coordinate transformations between the world frame, the robot base frame, and the end-effector frame. Since the proposed method explicitly models viewpoint changes and Manipulation policies in a structured manner, adapting it to dynamic base movements mainly requires consistent transformation handling, rather than redesigning the overall architecture.
\end{comment}

{\bf{Applicability to Different Embodiments}.} EgoReveal is not limited to robots with parallel grippers. Although this paper introduces EgoZero as the Manipulation policy, which assumes a parallel gripper configuration, overall EgoReveal does not depend on a specific end-effector design. Any manipulation policy tailored to a particular hardware embodiment can be integrated as a drop-in module. For example, for a five-fingered dexterous hand, a policy based on high-dimensional hand pose control or dexterous manipulation learning~\cite{bao2023dexart} could replace the parallel gripper policy, while keeping the active vision and occlusion assessment components unchanged. This modularity suggests that the framework can generalize across different robotic embodiments with minimal structural modification.

{\bf{Dealing with Complete Occlusion}.} Although EgoReveal addresses partial occlusion, EgoReveal can be extended to more challenging cases such as complete occlusion or situations where the target object is initially outside the camera's field of view. By introducing an additional stage for estimating or searching for the initial arm and camera configuration, the system can first actively explore the environment to locate the target object. Prior work on active object search~\cite{8976257} can be incorporated to identify a viewpoint where the object becomes at least partially visible. This viewpoint can then serve as the initial frame for our occlusion-aware manipulation pipeline. 

Despite these promising extensions, several limitations and challenges remain. Most notably, our module currently relies on an input reference image to determine whether an object is still occluded. This is a strong assumption that limits the broader applicability of the proposed module. Moreover, ORA utilizes color matching in the HSV color space for occlusion detection, which introduces further assumptions, such as the necessity for consistent lighting conditions and reliable color separability between objects.
% Addressing these algorithmic dependencies is a critical area for improvement.

As future work, we plan to expand our evaluation to include more diverse occlusion scenarios, broader object categories, and more complex manipulation tasks, supported by a larger and more diverse demonstration dataset. Furthermore, because our current reliance on scenario-specific policies limits scalability, a key direction is developing unified, cross-task Active Vision and Manipulation policies within a single model.
% Achieving this cross-task generalization is essential for deploying EgoReveal in unpredictable real-world environments.
% Despite these promising extensions, several challenges remain. As future work, we plan to extend our evaluation beyond the test scenarios considered in this study to include more diverse occlusion scenarios, a wider range of occluding objects, and various target objects. To support this extension, we intend to collect a larger and more diverse set of demonstration data. Furthermore, in the current implementation, separate policies are trained for each scenario. While this design simplifies learning, it limits scalability. Therefore, one possible future direction is to investigate the learning of a unified Active Vision policy and Manipulation policy capable of handling multiple occlusion patterns and task variations within a single model. Achieving such generalization would move EgoReveal closer to practical deployment in real-world environments, where occlusion patterns and object arrangements are highly diverse and unpredictable.
% モバイルmanipulater（台座ごと動く） --> 結局座標間の変換ができればいいだけなので適用可能って書く
% out of viewの時の対処（target objectの探索手法との組み合わせ）
% 5本指になったらどうか（egozeroじゃないフレームワークをもってきたらできる）

\section{CONCLUSIONS}
This paper addresses a novel task in which a robot learns to resolve partial occlusions and perform object manipulation from easily obtainable human demonstration data. To this end, we propose EgoReveal, a framework composed of three components: an Active Vision policy for resolving occlusions, a Manipulation policy for executing object interaction, and an Occlusion Resolution Assessment Module that determines when to switch between the two policies.
Experimental results demonstrate that the proposed method achieves higher task success rates compared to the baseline approach, validating the effectiveness of EgoReveal.
We believe that this work represents a meaningful step toward robot learning systems capable of addressing more complex real-world manipulation problems.

% We present a pipeline for learning Active Vision and Manipulation from human-centric data acquired via smart glasses, with subsequent deployment onto robotic systems. This methodology enables the efficient acquisition of control policies that demonstrate robustness against environmental challenges, such as occlusions. Our framework provides a scalable trajectory for future research in the field of Active Vision.

%%%%%%%%%%%%%%%%%%%%%%%%%%%%%%%%%%%%%%%%%%%%%%%%%%%%%%%%%%%%%%%%%%%%%%%%%%%%%%%%

%%%%%%%%%%%%%%%%%%%%%%%%%%%%%%%%%%%%%%%%%%%%%%%%%%%%%%%%%%%%%%%%%%%%%%%%%%%%%%%%

%%%%%%%%%%%%%%%%%%%%%%%%%%%%%%%%%%%%%%%%%%%%%%%%%%%%%%%%%%%%%%%%%%%%%%%%%%%%%%%%
% \section*{APPENDIX}
% \section*{ACKNOWLEDGMENT}

%%%%%%%%%%%%%%%%%%%%%%%%%%%%%%%%%%%%%%%%%%%%%%%%%%%%%%%%%%%%%%%%%%%%%%%%%%%%%%%%

\bibliographystyle{IEEEtran}
\bibliography{reference}

\end{document}